# Advancements in Weed Mapping: A Systematic Review

Mohammad Jahanbakht[123*], Alex Olsen[4], Ross Marchant[4], Emilie Fillols[5], and Mostafa Rahimi Azghadi[123*]

[1] College of Science and Engineering, James Cook University, Douglas, QLD 4814, Australia

[2] ARC Training Centre in Plant Biosecurity, James Cook University, Australia

[3] Agriculture Technology and Adoption Centre, James Cook University, Australia

[4] InFarm Pty Ltd, Goondiwindi, QLD 4390, Australia

[5] Sugar Research Australia, Brisbane 4000, QLD, Australia

[*]Corresponding Authors: Mohammad Jahanbakht (mohammad.jahanbakht@jcu.edu.au) and Mostafa Rahimi Azghadi (mostafa.rahimiazghadi@jcu.edu.au)

**Abstract** – Weed mapping plays a critical role in precision management by providing accurate and timely data on weed distribution, enabling targeted control and reduced herbicide use. This minimizes environmental impacts, supports sustainable land management, and improves outcomes across agricultural and natural environments. Recent advances in weed mapping leverage ground-vehicle Red Green Blue (RGB) cameras, satellite and drone-based remote sensing combined with sensors such as spectral, Near Infra-Red (NIR), and thermal cameras. The resulting data are processed using advanced techniques including big data analytics and machine learning, significantly improving the spatial and temporal resolution of weed maps and enabling site-specific management decisions. Despite a growing body of research in this domain, there is a lack of comprehensive literature reviews specifically focused on weed mapping. In particular, the absence of a structured analysis spanning the entire mapping pipeline, from data acquisition to processing techniques and mapping tools, limits progress in the field. This review addresses these gaps by systematically examining state-of-the-art methods in data acquisition (sensor and platform technologies), data processing (including annotation and modelling), and mapping techniques (such as spatiotemporal analysis and decision support tools). Following PRISMA guidelines, we critically evaluate and synthesize key findings from the literature to provide a holistic understanding of the weed mapping landscape. This review serves as a foundational reference to guide future research and support the development of efficient, scalable, and sustainable weed management systems.

## I. Introduction

Invasive weeds present a major threat to both agricultural productivity and environmental sustainability globally. Their spread leads to considerable economic losses by competing with the crop and reducing crop yields, disrupting harvesting, and promoting insect pests and diseases [1, 2]. As the agricultural sector contends with these impacts, effective weed management strategies are crucial to support sustainable food production and protect natural ecosystems.

One such strategy involves integrated precision approaches, leveraging technological advancements to enhance weed mapping and control. In this regard, precision agriculture represents a paradigm shift from traditional farming systems, utilizing data-driven methodologies to optimize resource utilization and improve productivity [3]. It integrates modern electronic sensors, including those for monitoring environmental parameters, navigation, visual and spectral imaging, and mapping, with advanced processing techniques such as machine learning algorithms to assess conditions across the field. This enables informed, site-specific decision-making tailored to both farm and environmental conditions [4].

Recent advancements in electronic sensors, including those employed in remote sensing technologies have revolutionized agricultural and environmental monitoring, offering enhanced capabilities for weed mapping. Remote sensing platforms, including satellites [5] and Unmanned Aerial Vehicles (UAVs) [6], provide large-scale spatial and sometimes temporal data that enable precise weed detection and management. In addition, ground-based machinery can be used to collect weed data at high spatial and temporal resolution [7].

Once weed data is collected, it must be processed and analyzed for weed detection and classification. This is typically achieved using machine learning models capable of handling complex spectral and spatial data, allowing for the reliable identification of weed species even in diverse and heterogeneous agricultural landscapes [8]. The integration of advanced remote sensing technologies, such as hyperspectral and multispectral imaging, further enhances these models by providing rich spectral information that helps

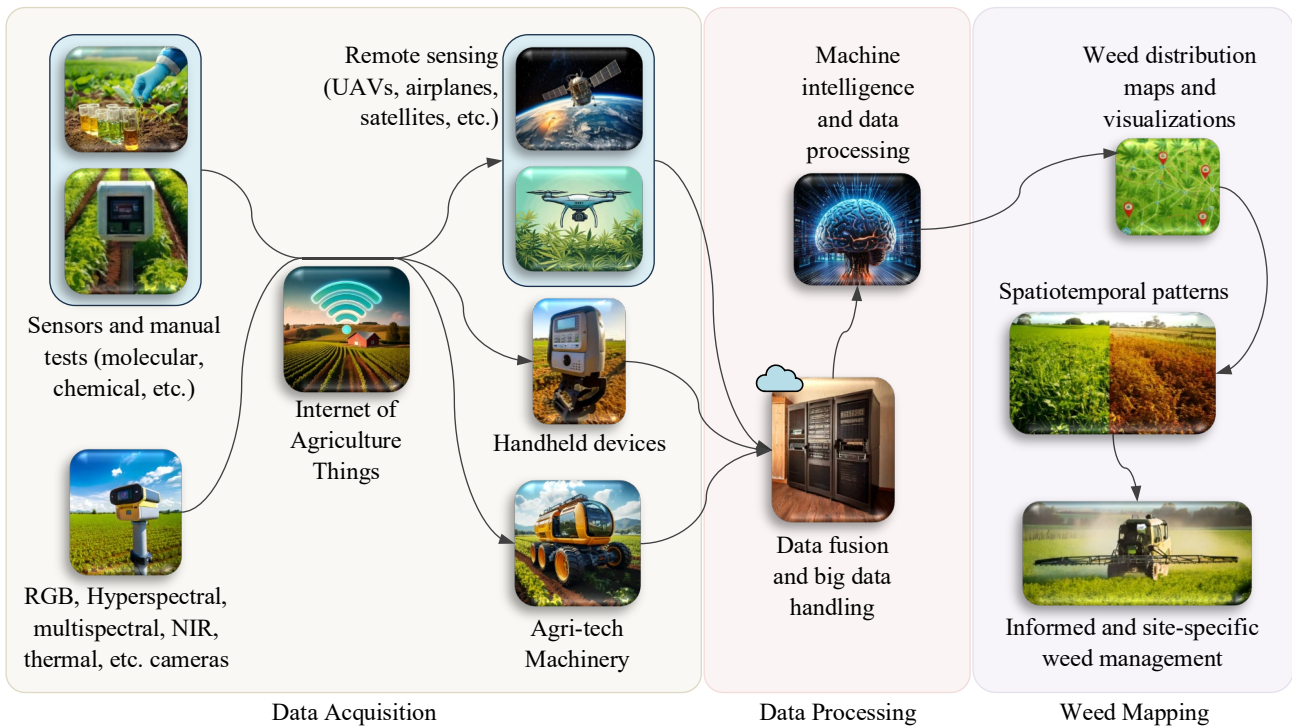

**Figure 1**: This paper covers the full weed mapping pipeline as illustrated here, from data acquisition to advanced data processing, to visualization dashboards and spatiotemporal models that support informed farm management.

distinguish plant species based on their unique spectral signatures [9].

After invasive weed detection with advanced machine learning and analytical models, analyzing the spatiotemporal patterns of weed spread is crucial for understanding their ecological impact and predicting future invasions. This step enables more effective, data-driven management strategies by identifying high-risk areas and optimal intervention times. Recent advancements in spatiotemporal weed pattern analysis have significantly enhanced our understanding of weed dynamics across agricultural landscapes. Innovative technologies, such as timeseries forecasting and photogrammetry, now enable the creation of detailed multi-dimensional models of weed populations, capturing variations in plant height, volume, and canopy structure over time [10]. These models facilitate the generation of high-resolution spatiotemporal maps, allowing for precise monitoring of weed distribution and growth patterns.

To explore these novel advancements in weed mapping technologies, we will systematically review the latest developments in the area, highlighting their potential to enhance efficiency and sustainability. By examining current research, technological innovations, and practical applications, we aim to provide insights into how precision management can be harnessed to improve weed control strategies. This study's primary contributions include:

- Filling a significant gap in the literature by developing a comprehensive systematic review focused solely on weed mapping serving as a central reference point for future studies and applications.

- Providing a detailed and structured synthesis of modern data acquisition tools and technologies used in weed mapping, offering clarity on their applicability and limitations.

- Reviewing and synthesizing current data processing methodologies, including big data handling, annotation strategies, machine learning, deep learning, and edge computing, by highlighting their strengths, challenges, and practical implications.

- Exploring commonly used weed mapping tools, spatial and temporal pattern modelling approaches, and their integration into decision support systems, providing valuable insights for operational deployment.

- Delivering cross-sectoral actionable insights for agricultural technology developers, policymakers, and farm managers aiming to implement data-driven and environmentally sustainable weed control strategies.

- Identifying key research and development opportunities, encouraging interdisciplinary innovation in weed mapping technologies and offering a future-oriented roadmap.

The rest of the paper is structured as follows. Section II presents our Preferred Reporting Items for Systematic Reviews and Meta-Analyses (PRISMA)-based systematic review methodology, outlining the literature selection, screening, and analysis process. The overall flow of technical discussions is summarized in Figure 1, covering acquisition, processing, and mapping. Section III discusses data acquisition tools used in weed mapping, including a review of imaging sensors, agricultural machinery, and remote sensing platforms such as drones and satellites. Section IV explores weed data processing techniques and technologies, covering key aspects such as big data handling, data annotation, deep learning models, and edge computing solutions. Section V focuses on weed mapping technologies, discussing spatiotemporal weed patterns, the impact of farm management practices, and commonly used mapping tools and decision-support systems. Finally, Section VI outlines future research directions and technological innovations needed to advance the field of precision weed management.

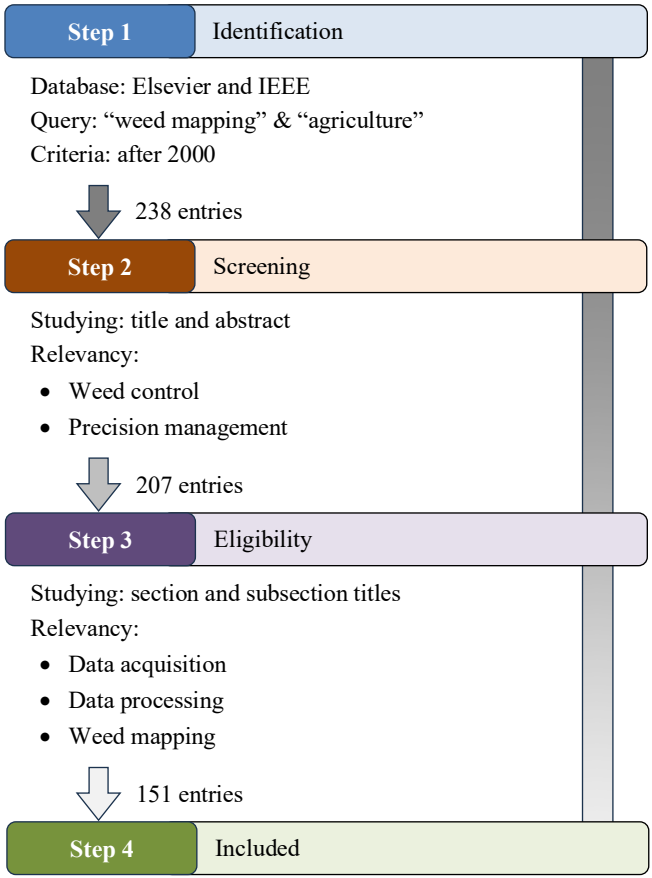

**Figure 2**: The literature selection flowchart through consecutive inclusion/exclusion steps that follows the PRISMA guidelines.

## II. Survey Methodology

This systematic review was conducted following the PRISMA guidelines, ensuring a transparent, methodical, and replicable approach to synthesizing the literature on weed mapping in agriculture. The use of PRISMA enhances the scientific quality of the study by reducing bias through comprehensive and structured literature searches, while providing a clear account of the procedures undertaken throughout the review process. The methodology was carried out in three main phases: identification, screening, and eligibility, as illustrated in Figure 2.

In the identification phase, we performed an initial search using the keywords "weed mapping" and "agriculture" across two major academic databases: Elsevier and IEEE Xplore. This search was conducted in April 2025 and limited to publications released after the year 2000. A total of 238 articles (including 222 papers from Elsevier and 16 papers from IEEE Xplore) were retrieved and archived locally for further analysis.

The screening phase involved reviewing the titles and abstracts of all 238 identified records. To proceed to the next stage, studies were required to meet two criteria: (a) the research addressed weed control topics, and (b) the application context was within precision weed management. After applying these filters, 207 articles were retained.

In the final eligibility phase, the remaining 207 articles were subjected to a more in-depth evaluation. We examined section and subsection titles and partially skimmed the main text of each article to ensure relevance. Studies were excluded if they (a) did not pertain to modern agricultural methods in weed management, or (b) failed to address at least one of our three core focus areas: data acquisition, data processing, and weed mapping. It is worth noting that some of the studies were included in more than one focus area. Ultimately, 151 publications met the eligibility criteria and were included in this review.

These 151 studies comprise 135 journal articles, 8 government, web, book, or thesis reports, and 8 conference papers. Their publication years span from 2003 to 2025. A detailed breakdown of these studies based on publication type, publication year, and conceptual focus is presented in Figure 3.

## III. Data Acquisition

Traditional methods for monitoring incursions of invasive weeds are often labor-intensive, time-consuming, expensive, and rarely fully effective [2]. Advancements depend on the development of efficient and scalable data acquisition and processing technologies. Innovative tools, including cameras and sensor-based systems, must be capable of addressing the dynamic characteristics of weeds and the vast scale of agricultural landscapes to enable effective detection and tracking of infestations. This section will explore modern approaches to data acquisition in weed surveillance.

**Data Capture Modalities**

The integration of advanced imaging technologies in agriculture has significantly enhanced weed mapping and crop monitoring. These technologies leverage different parts of the electromagnetic spectrum in Figure 4a to capture vital information about crops, soil, and surrounding vegetation. The following subsections discuss various imaging modalities and their applications in agricultural practices, particularly in weed management.

*X-ray*

X-ray technology is widely utilized in agriculture for detecting contaminants in food packaging and assessing the quality of agricultural products. X-rays reveal spatial information and acquire three-dimensional data, making them effective for detecting density variations in varied materials. Most agricultural applications employ soft X-rays, which have been extensively used for studying crops, soil, grains, tree nuts, and fruits [11]. Soft X-ray technology with low energy and longer wavelength (compared to hard X-rays) allows for detailed visualization of internal structures in thin-film materials, making it a valuable tool for quality assessment in agricultural products.

In weed management, soft X-ray imaging plays a crucial role in seed inspection, commonly known as seed radiography [12]. This technique helps identify and remove weed seeds from agricultural good seed batches by analyzing their internal structures before planting. By sorting out undesirable weed seeds, farmers can reduce the risk of weed infestations, leading to improved crop productivity and quality.

*Visible RGB*

RGB imaging relies on visible light to capture high-resolution images of crops and weeds. The quality of image acquisition is dependent on two primary components: illumination sources and camera systems. The choice of illumination significantly influences the ability to extract texture, shape, and color features of agricultural objects.

In a study by Raja *et al*. [13], multiple illumination techniques are successfully utilized to enhance the accuracy of weed classification in controlled-light imaging chambers. These chambers were designed to minimize interference from natural light and to capture high-resolution images under uniform illumination, white balance calibration, and controlled exposure time, ensuring clear visualization of crops and weeds. This integration of illumination techniques has significantly enhanced the reliability of the classification algorithm, achieving a lettuce crop detection accuracy of 99.75% and correctly identifying 98.11% of sprayable weeds.

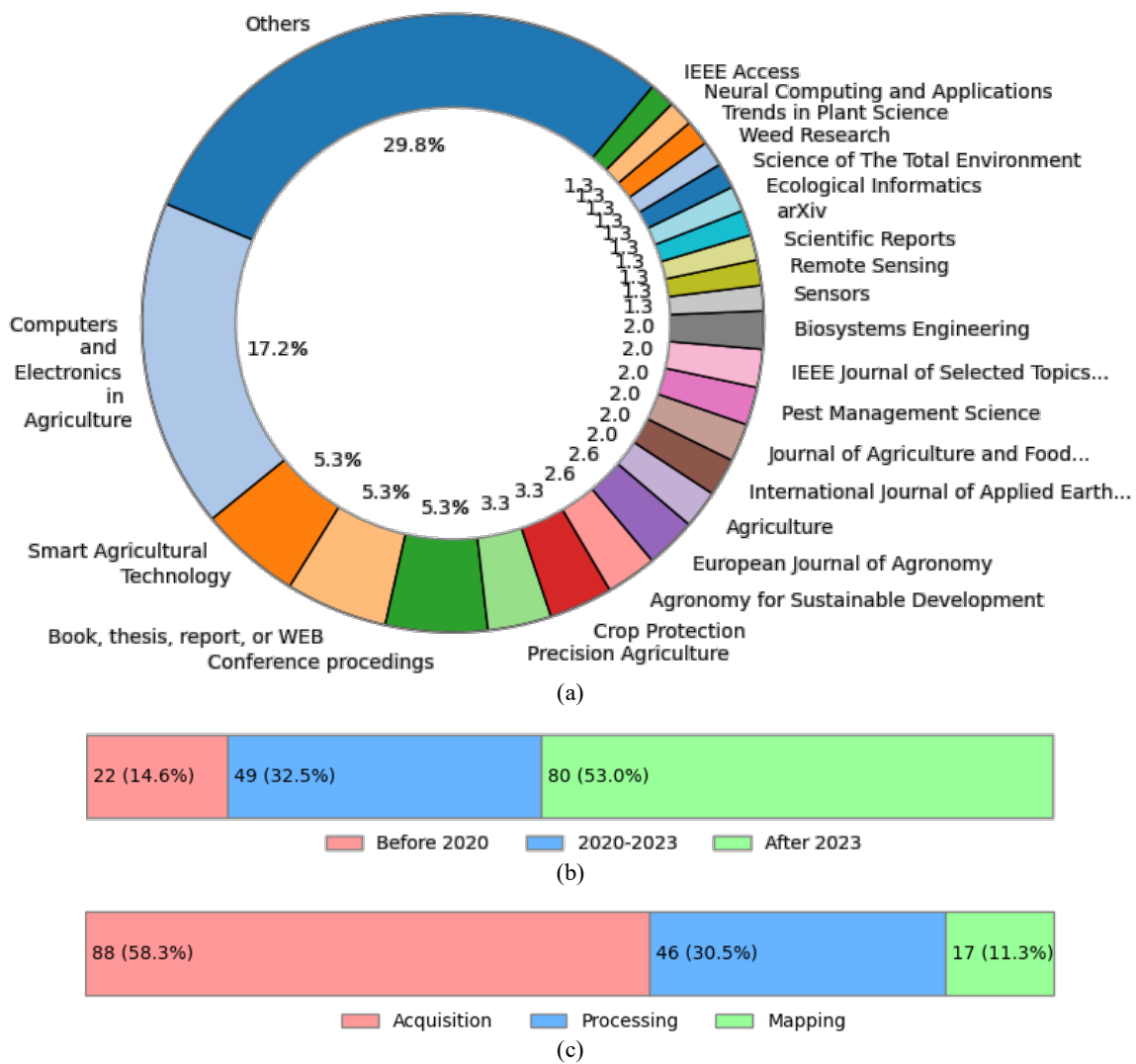

**Figure 3**: Distribution of the surveyed articles over (a) publication venues, i.e., journal names, conferences, official reports, websites, etc., (b) publication years, and (c) conceptual topics of this review, i.e., data acquisition, data processing, and weed mapping.

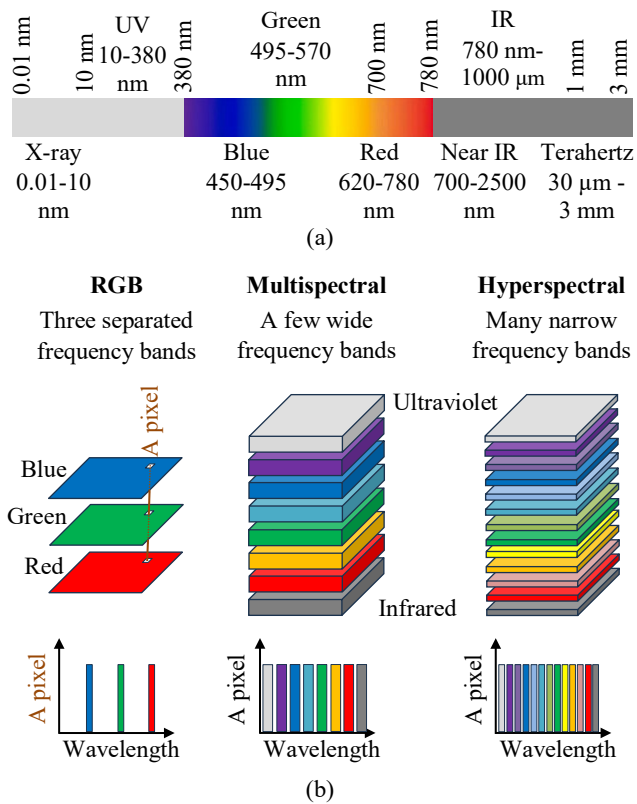

**Figure 4**: (a) Frequency spectrum and bandwidths in agricultural applications and (b) Conceptualization of RGB, multispectral, and hyperspectral imaging techniques.

In another work addressing the illumination problem, FieldNet was proposed as a real-time deep learning framework for shadow removal in outdoor environments. By eliminating shadows without needing shadow masks at inference, FieldNet improves image consistency under varying lighting, enhancing weed detection accuracy in field robotics [14].

Cameras, the other major component of RGB imaging, include monocular and binocular configurations. Monocular cameras provide cost-effective 2D imaging, while binocular stereo vision (stereoscopic) systems generate 3D visual representations by capturing depth information. The former is the most common scenario, while the latter is particularly useful for measuring object dimensions and detecting plant structures, making it a valuable tool for weed identification applications.

Binocular cameras are successfully utilized in [15] for weed detection in rice fields, significantly improving classification accuracy compared to conventional single-source cameras. By capturing stereoscopic video data and incorporating 3D depth perception under controlled light conditions, the study leveraged a computer vision system that achieved 96.95% weed classification accuracy. This helps distinguish between similar-looking plants more effectively than single-camera systems.

*Spectral*

Spectral remote sensors have transformed the way we collect and analyze data on various weed species across different environments. These advanced sensors capture detailed spectral reflectance information from target plants, supporting agricultural applications such as weed identification, crop yield estimation, and disease monitoring. Spectral imaging includes multispectral and hyperspectral techniques [16]. Figure 4b compares these techniques with each other and with the RGB imaging method.

Hyperspectral imaging captures hundreds of narrow spectral bands, providing extensive spectral information for each pixel in an image. This capability is highly beneficial for material identification, agriculture, mineralogy, and medical imaging. In agriculture, hyperspectral imaging is used to classify weeds, detect disease and pest, monitor plant health, and plant stressors [16]. However, hyperspectral imaging systems are costly and require complex data modeling and processing. The high spectral resolution can also limit real-time applications due to computational challenges. Despite these drawbacks, hyperspectral imaging remains an essential tool for precise weed mapping and vegetation monitoring.

On the other hand, multispectral imaging captures a limited number of broader spectral bands, making it a more computationally efficient alternative to hyperspectral imaging. It is commonly used in remote sensing applications and environmental monitoring. Compared to hyperspectral imaging, multispectral imaging offers a balance between computational efficiency and spectral information, making it a widely used approach for weed detection and agricultural mapping [9].

Table 1 presents various weed species commonly found in agricultural fields, detailing their names, associated crop environments, and the type of vision technology used for their detection. The listed weeds span multiple crops, including sugarcane, wheat, maize, soybean, and barley, highlighting their widespread impact on global agriculture. This highlights the effective use of RGB and spectral imaging sensors in advanced weed detection studies in literature.

Table 1: RGB and spectral imaging technologies reported in the latest weed control applications.

| | | | | | |
|---|---|---|---|---|---|
| RGB | Weed | *Amaranthus blitoides* 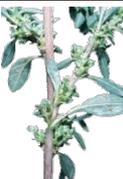 | *Amaranthus tuberculatus* 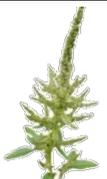 | *Chromolaena odorata* 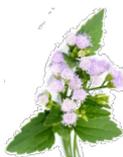 | *Cirsium arvense* 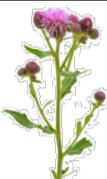 |
| | Crop | Maize [17] | Black bean, canola, corn, flax, soybean, and sugar beets [18] | Crop of tropical climate regions [19] | Wheat [20] and barley [21] |
| | Weed | *Cynodon dactylon* 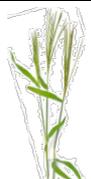 | *Ambrosia artemisiifolia* 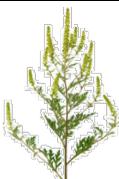 | *Sorghum halepense* 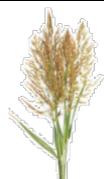 | *Tussilago farfara* 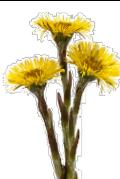 |
| | Crop | Vine [22] | Black bean, canola, corn, flax, soybean, and sugar beets [18] | Maize [17] | Barley [21] |
| Spectral | Weed | *Ageratum conyzoides* 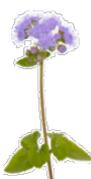 | *Avena sterilis* 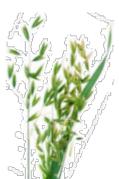 | *Amaranthus palmeri* 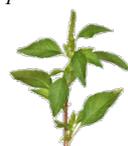 | *Bassia scoparia* 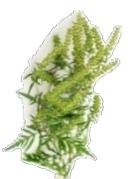 |
| | Crop | Sugarcane [23] | Wheat [24] | Corn, soybeans, and cotton [25] | Barley, corn, dry pea, garbanzo, lentils, pinto bean, safflower, and sugar beet [26] |
| | Weed | *Chenopodium album* 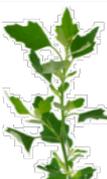 | *Cirsium arvense* 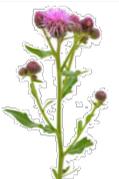 | *Commelina benghalensis* 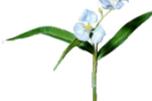 | *Conyza canadensis* 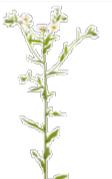 |
| | Crop | Maize [27] | Maize and sugar beet [27] | Sugarcane [23] | Barley, corn, dry pea, garbanzo, lentils, pinto bean, safflower, and sugar beet [26] |
| | Weed | *Crotalaria juncea* 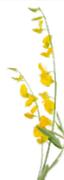 | *Fallopia convolvulus* 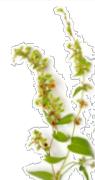 | *Ipomoea hederifolia* and *Ipomoea purpurea* 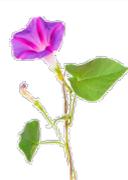 | *Lolium multiflorum* and *Lolium rigidum* 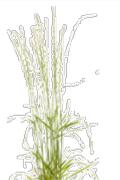 |
| | Crop | Sugarcane [23] | Sugar beet [27] | Sugarcane [23] | Sugar beet [27] and wheat [24] |
| | Weed | *Megathyrsus maximus* 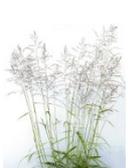 | *Phalaris brachystachys* 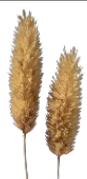 | *Sorghum halepense* 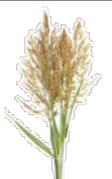 | *Urochloa brizantha* 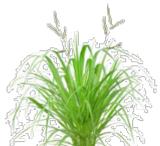 |
| | Crop | Sugarcane [23] | Wheat [24] | Maize [28] | Sugarcane [23] |

### NIR and Thermal

Both Near-Infrared (NIR) and thermal cameras use infrared radiation but differ in their detected wavelength ranges. NIR imaging detects reflected light in the 700–2500 nm range, while thermal imaging captures emitted heat in the 3–14 μm range. These technologies offer critical insights into plant health, transpiration rates, and water potential.

In weed control applications, NIR and thermal cameras can be used in temperature-based differentiation [29].

Weeds often exhibit different thermal properties compared to crops due to variations in water content, leaf structure, and metabolic activity. For example, weeds, competing with crops for resources like water, may have different water content levels, influencing their thermal behavior [30]. Furthermore, leaf structures of crops and weeds, including area and thickness, as well as their metabolic activities (reflected in processes like photosynthesis and respiration) contribute to a plant's energy balance and thus its thermal properties [31].

*Terahertz*

Terahertz (THz) imaging is an emerging technology used for detecting small unwanted objects in agricultural environments, such as pests, worms, or foreign bodies in crop yields. This technique employs orthogonally polarized terahertz waves to enhance detection accuracy in various agricultural settings, including conveyor belts and land surveying vehicles.

As THz waves interact differently with plant tissues (based on their water content, chemical composition, and cellular structure) a precise differentiation can be detected between weed species and crops, even in dense vegetation. This differentiation can help identify weeds at early growth stages, even before visible differences appear. THz imaging can also be combined with machine learning algorithms to improve weed classification accuracy, helping farmers and researchers develop targeted weed control strategies [32].

**Data Capture Equipment**

*Agri-tech Machinery*

The application of imaging technologies (discussed in the previous subsection) can benefit from Agri-tech machinery that enables improved spatial resolution, greater coverage, and enhanced temporal flexibility. Among the most appealing approaches are All-Terrain Vehicles (ATVs) and autonomous robotic systems, which provide flexible and responsive solutions to farming operations. These intelligent machines, equipped with state-of-the-art navigation systems, sensors, and AI-driven automation, enable efficient and precise location-specific agricultural tasks, even in challenging terrains [1].

ATVs and agricultural robots enhance productivity by facilitating various farming operations, ranging from sowing and field monitoring to weed control, and harvesting. By leveraging machine learning and vision systems, these autonomous machines optimize resource use while reducing crop damage and soil compaction. Especially in field monitoring and health assessment applications, ATVs and robotic systems can process real-time data collected from sensors and imaging systems to evaluate key indicators such as leaf color, biomass, and disease symptoms [3].

In advanced weed control systems, these machineries employ sophisticated image recognition and classification models to differentiate between crops and weeds [33]. By leveraging real-time sensor data, they execute precise and automated physical weed removal or targeted herbicide applications, reducing the reliance on broad-spectrum chemical treatments. This approach enhances crop health, minimizes agrochemical overspray, and lowers environmental risks associated with traditional weed management strategies [34]. The collection of studies in Table 2 explores various machine vision and AI-based techniques for precision agriculture, focusing on weed and pest detection, automated spraying, and robotic weeding. Commercial solutions (e.g., Trimble's Bilberry, John Deere's See & Spray, and GreenEye) are excluded, as the table focuses exclusively on research-based literature. Besides, some of the listed machines, for example [35], have not been directly used for weed data collection, detection, and management, they provide examples of other machinery and sensors applicable to weed data collection.

*Remote Sensing*

Remote sensing involves collecting physical information about an object without direct contact [36]. This plays a crucial role in precision agriculture by enabling crops and farmlands monitoring from varying distances. Technologies such as UAVs and satellites provide superior support for agricultural applications, assisting in crop scouting, yield estimation, precise agrochemical application, and weed control.

A major challenge in remote sensing for precision agriculture lies in spatial, temporal, spectral, and radiometric image resolutions. Spatial resolution determines image detail based on pixel density, with Ground Sampling Distance (GSD) serving as a key metric. Temporal resolution reflects how frequently a location is imaged, while spectral resolution influences the ability to differentiate objects based on narrow frequency bandwidths. Radiometric resolution defines a sensor's capacity to capture subtle energy variations, affecting the precision of plant identification [37].

High resolutions across these categories are critical for distinguishing crops and weeds, particularly in diverse agricultural landscapes. UAVs have emerged as a dominant technology in precision agriculture due to their ability to capture high-resolution images with RGB, multispectral, and hyperspectral sensors, perform low-altitude maneuvers, access difficult terrain, and operate at a lower cost while being less affected by weather conditions compared to satellites. However, UAVs face limitations, including restricted area coverage, payload constraints, regulatory challenges, and the need for skilled operators [37]. Both the drone and satellite applications in weed mapping are studied in more detail here.

**Table 2**: The application type and operation method of research-based smart Agri-tech machinery and robots that have been or can be used in industrial weed management systems.

| Appl. | Description | | Appl. | Description | |
|---|---|---|---|---|---|
| Robot [35] (for greenhouses) | • Detecting thrips in strawberry greenhouses<br>• Traditional machine learning (i.e., SVM)<br>• Utilizing region and color indices to classify pests, with different kernel functions applied for improved accuracy | 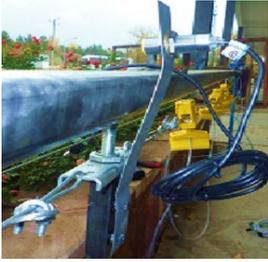 | ATV [42] (for orchard spraying) | • Precise and automatic spraying system for peach orchards<br>• Detects the leaf wall area and plans spraying paths based on region of interest (except in areas with row gaps)<br>• Traditional machine learning (i.e., depth features and Otsu)<br>• Using color-depth vision | 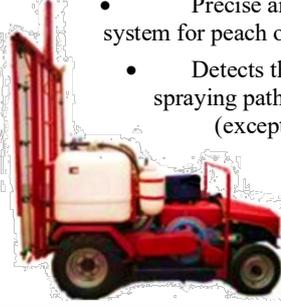 |
| Robot [38] (general purpose) | • Air-blast spraying of citrus orchards<br>• Low-cost smart sensing system using LiDAR<br>• Deep learning (i.e., YOLOv3 with Resnet50 backbone)<br>• Classifying trees, estimating tree heights, counting fruits, and enabling precise nozzle control for targeted spraying | 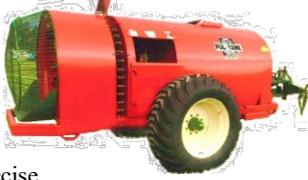 | ATV [43] (for orchard spraying) | • Intelligent spraying system for pear orchards<br>• Deep learning (i.e., SegNet)<br>• Semantic segmentation of fruit trees<br>• Integrating depth data from an RGB-D camera to avoid detecting background trees and controls nozzles based on tree coverage in image zones | 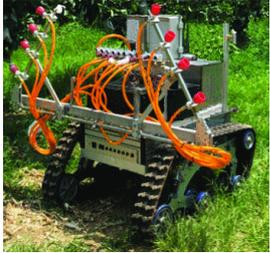 |
| ATV [39] (for row farming) | • Mechanical weeding machine for precise weed removal in cultivation aisles<br>• A modular weeder with an inverted pyramid-shaped tool efficiently shovels weed out without use of herbicide<br>• Deep learning (i.e., convolutional neural network)<br>• Accurately identifying/detecting weeds | 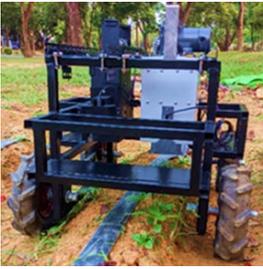 | Robot [33] (called BonnBot) | • Weed detection in sugar-beet<br>• Integrating ecological considerations into precision weeding robots<br>• Rolling-view observation model to improve weeding performance and oversee diverse weed distributions<br>• Deep learning (i.e., Mask-RCNN) | 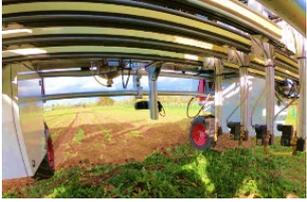 |
| ATV [40] (for non-row farming) | • A weeding robot in corn fields<br>• Equipped with a quadratic traversal algorithm for guiding around the identified corn plants<br>• Deep learning (i.e., Faster R-CNN)<br>• real-time image processing on edge and shortest 3D path calculation, based on plant contours and depth cameras. | 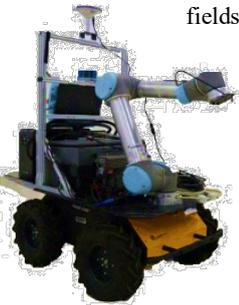 | Robot [44] (for hydroponic farms) | • Weed detection in celery houses<br>• Real-time robotic weed control in dense vegetable fields by treating celery plants with Rhodamine B to create machine-readable fluorescent signals<br>• Traditional machine learning (i.e., segmentation by color features)<br>• Custom illumination system (spectral fluorescence imaging) for precise differentiation of crops from weeds and accurate stem localization | 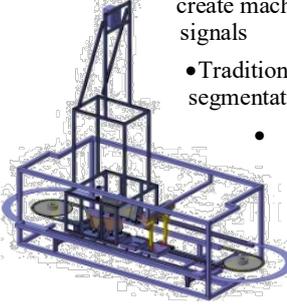 |
| Robot [41] (general purpose) | • Weed detection in corn fields<br>• Traditional machine learning (i.e., green features and Otsu)<br>• Identifying/ segmenting and positioning corn, weeds, and land profile to improve weed removal efficiency | 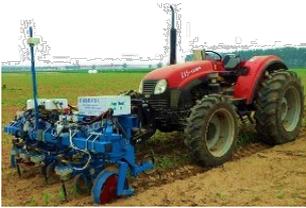 | Robot [34] (for row farming) | • Effective weed detection and control in strawberries<br>• Autonomous laser weeding robot with minimal seedling damage<br>• Deep learning (i.e., YOLOv8)<br>• Detects strawberry seedlings, weeds, drip irrigation pipes, and weed growth points in real-time. | 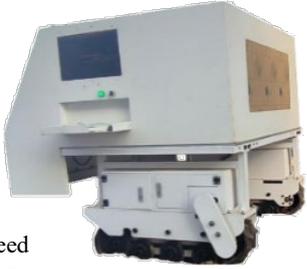 |

**Table 3**: Common satellites in large-scale agricultural applications that have been or can be applied in weed mapping. Click the blue hyperlinks to access the satellite webpage.

| Satellite | Resolution | Applications |
|---|---|---|
| EnMap<br>DLR (Germany)<br>Since 2022 | Spatial: 30 m<br>Swath: 30 km<br>Temporal: 27 days | HS (420-2450 nm)<br>Crop mapping [45]<br>Plant health [46] |
| EO-1 (Hyperion)<br>NASA (USA)<br>Since 2000 | Spatial: 30 m<br>Swath: 7.7 km<br>Temporal: 16 days | HS (357-2576 nm)<br>Crop mapping [47, 48]<br>Plant health [49] |
| Ikonos<br>ESA (Europe)<br>Since 1999 | Spatial: 1 m<br>Swath: 11.3 km<br>Temporal: 3 days | MS (4 bands)<br>Crop mapping [50] |
| KOMPSAT-3<br>SI (Korea)<br>Since 2012 | Spatial: 2 m<br>Swath: 15 km<br>Temporal: 1.4 days | MS (2 bands)<br>Land cover [51] |
| Landsat 8/9<br>NASA (USA)<br>Since 2013 | Spatial: 30 m<br>Swath: 185 km<br>Temporal: 16 days | MS (11 bands)<br>Crop mapping [47]<br>Weed mapping [52] |
| PlanetScope<br>ESA (Europe)<br>Since 2016 | Spatial: 3 m<br>Swath: 25 km<br>Temporal: 1 day | MS (4 bands)<br>Crop mapping [53]<br>Weed mapping [54] |
| Pléiades<br>CNES (France)<br>Since 2011 | Spatial: 0.5 m<br>Swath: 20 km<br>Temporal: 1 day | MS (6 bands)<br>Crop mapping [55] |
| Prisma<br>ASI (Italy)<br>Since 2019 | Spatial: 30 m<br>Swath: 30 km<br>Temporal: ~29 days | HS (400-2500 nm)<br>Land cover [56]<br>Crop mapping [57] |
| Proba 1<br>ESA (Europe)<br>Since 2001 | Spatial: 17 m<br>Swath: 15 km<br>Temporal: 7 days | HS (400-1300 nm)<br>Plant health [49] |
| Sentinel 1<br>ESA (Europe)<br>Since 2014 | Spatial: 10 m<br>Swath: 250 km<br>Temporal: 6-12 days | Radar (C-band)<br>Crop mapping [58]<br>Land cover [59] |
| Sentinel 2 A/B<br>ESA (Europe)<br>Since 2015 | Spatial: 10 m<br>Swath: 290 km<br>Temporal: 5 days | MS (13 bands)<br>Land cover [59, 60]<br>Weed mapping [61] |
| SPOT 6/7<br>ESA (Europe)<br>Since 2012 | Spatial: 6 m<br>Swath: 60 km<br>Temporal: 13 days | MS (5 bands)<br>Crop mapping [62]<br>Land cover [63] |
| WorldView-3<br>ESA (Europe)<br>Since 2014 | Spatial: 0.31 m<br>Swath: 13.1 km<br>Temporal: ≤1 day | MS (16 bands)<br>Crop mapping [64] |

MS: Multispectral, HS: Hyperspectral

***Satellites***: Before the emergence of drones, satellites were the primary platform for agricultural remote sensing due to their widespread availability and cost-effectiveness. Satellite-based remote sensing has played a crucial role in monitoring large-scale agricultural landscapes, providing valuable insights into crop health, soil conditions, and environmental factors. Satellites are equipped with a variety of sensors, including optical, multispectral, hyperspectral, radar, and thermal imaging technologies, making them versatile tools for precision agriculture [45].

One of the key advantages of satellite remote sensing is its ability to cover vast or inaccessible areas where traditional field-based data collection methods would be impractical. Several commercial and freely available satellites are equipped with image sensors. However, high-resolution commercial satellite images can be expensive, limiting access for smallholder farmers. A selection of available free and commercial satellites is summarized in Table 3.

Despite their advantages, satellites have inherent limitations. Cloud cover can obstruct their view, while atmospheric effects like scattering and absorption may distort the accuracy of the images they capture. As a result, cloud detection models and atmospheric correction techniques are required to adjust satellite radiation measurements and accurately interpret surface reflectance. Additionally, reflections from the surface or lower atmosphere may alter the true reflectance properties of agricultural materials, requiring further calibration [5].

Another challenge with widely used satellite systems, such as Landsat and Sentinel-2, is their relatively low spatial resolution (typically 10–30 meters), which restricts their ability to capture fine-scale agricultural variations. To address this, high-resolution commercial satellites, such as WorldView-3 and Ikonos, have gained popularity in recent years [50]. These satellites offer spatial resolutions as high as 1–3 meters, enabling detailed agricultural monitoring that medium-resolution satellites cannot achieve.

Beyond spatial resolution, commercial satellites often provide additional spectral bands and flexible revisit times. For instance, WorldView-3 includes shortwave infrared and red-edge bands, which improve the detection of crop residues and vegetation characteristics. Additionally, many commercial satellite services allow on-demand tasking, offering higher-frequency data acquisition for specific agricultural regions compared to freely available satellites with fixed revisit schedules [45].

As an example of high-resolution commercial satellite application, a noteworthy study has been conducted by Shendryk *et al.* [65], which focuses on mapping the spread of *Andropogon gayanus* (gamba grass). Gamba grass is an invasive pasture grass that is rapidly spreading through the tropical savannas of northern Australia, increasing fire intensity, and causing ecological damage. To effectively monitor and manage its spread, the researchers developed a machine learning model to ingest high-resolution WorldView-3 satellite imagery. The results demonstrated that under optimal conditions, gamba grass can be mapped from satellite imagery with an accuracy of up to 91%. Additionally, spectral indices derived from the imagery significantly improved detection accuracy compared to using raw spectral bands alone.

*Drones*: While satellites remain indispensable for large-scale and long-term agricultural monitoring, drones have revolutionized precision agriculture by offering ultra-high-resolution imagery with greater flexibility. Advances in sensory and imaging technologies, along with improvements in data processing techniques, continue to enhance the role of drone remote sensing in modern precision farming.

Drones, also known as Unmanned Aerial/Aircraft Systems (UAS), offer a cost-effective way to collect aerial data. Although they generate large volumes of data that demand substantial storage and processing, drones can enable farmers to increase productivity and make informed decisions through real-time aerial observation, early disease detection, targeted interventions, and improved agricultural sustainability. A list of drones that have been or can be used in weed mapping is presented in Table 4. Specifically, their capability to flexibly cover large areas and generate high-resolution images aids in identifying and managing weed patches [66].

Site-specific weed management using drones is gaining popularity [67, 68]. This approach involves precisely targeting weed control methods to individual weeds or weed patches, accounting for spatial variability and temporal dynamics rather than uniformly treating the entire field. Since weeds typically grow in clusters rather than being evenly distributed, site-specific management presents a significant opportunity for reducing herbicide use while maintaining effective weed control [69].

The study in [67] explores a site-specific weed control approach in corn fields using a UAV to map weed distribution, generate a prescription map, and selectively spray using a commercial sprayer. A Crop Row Identification algorithm was developed to detect and remove corn rows from drone imagery, classifying remaining vegetation as weeds. A grid-based prescription map guided herbicide application, ensuring only grids with detected weeds were sprayed. This method reduced herbicide application by 26.2% compared to conventional practices, demonstrating the potential for reducing chemical use in corn production while maintaining effective weed control.

**Table 4**: The most common drone products in field monitoring that have been or can be applied in weed mapping.

| Drone Type | Sensor | Applications |
|---|---|---|
| Batmap, Nuvem Fixed wing 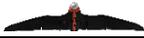 | RGB | Plant detection [70] |
| DJI, Matrice 100/300/600 Quadcopter 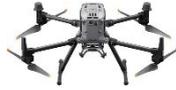 | RGB | Yield estimation [71] field mapping for spraying [72], and plant growth analyses [73] |
| | Multi-spectral | Pest infestation mapping [74], disease detection [75], and data fusion in agriculture [76] |
| DJI, Mavic 2/3/Air/Pro Quadcopter 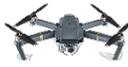 | RGB | Plant growth analyses [73], plant detection [77, 78], pest detection [79], and weed detection [19] |
| DJI, Phantom 3/4 Quadcopter 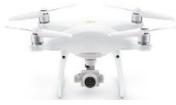 | RGB | Weed detection [80, 81], crop detection [78, 82], and field mapping [83] |
| | Multi-spectral | Disease classification [84, 85] and plant health monitoring [86] |
| DJI, s1000 Octocopter 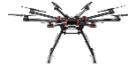 | Multi-spectral | Plant detection [87] and plant health monitoring [88] |
| PFT, Firefly 6 Fixed wing 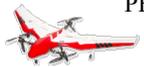 | RGB | Field mapping [89] |
| Horus, Aeronaves Fixed wing 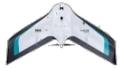 | RGB | Weed segmentation [90] |
| Microdrones, md4 Quadcopter 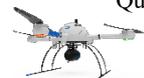 | RGB | Annotated weed imagery dataset [91] and weed segmentation [92] |
| Parrot, Anafi Quadcopter 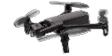 | RGB | Disease detection [75] |
| Parrot, Bluegrass Quadcopter 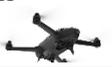 | Multi-spectral | Plant health monitoring [93] and field mapping [94] |
| Quantum systems, F90+ Fixed wing 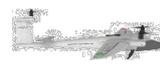 | Thermal | Disease classification [84] |
| SenseFly, Ebee Fixed wing 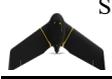 | RGB | NDVI greenness estimation [95] |

To address the gap in sensor performance evaluation, Betitame *et al.* [96] compared the performance of UAV-mounted RGB and multispectral sensors in distinguishing crops, broadleaf weeds, and grasses in soybean fields. Using traditional classification algorithms and object-based image analysis in ArcGIS

Pro, results showed that the RGB sensor achieved 93.8% accuracy, while the multispectral sensor had a similar accuracy of 93.4%. The RGB sensor performed better at minimizing misclassifications and was particularly effective in detecting grass, while the multispectral sensor excelled in estimating total crop area due to its broader spectral range. Both sensors effectively classified background regions. Given the comparable performance, the less expensive RGB sensor may be more suitable for cost-effective precision agriculture applications.

*Ecballium elaterium* (a.k.a., squirting cucumber) is a difficult-to-control weed in non-tillage olive groves, infesting inter-row cover crops. Given its patchy distribution, site-specific control strategies can be effective. The study conducted in [68] developed a UAV-based methodology to detect and map *E. elaterium* infestations using RGB imagery. Conducted in two super-intensive olive orchards, UAV flights captured images in May (with multiple weed species) and September (when *E. elaterium* was the sole weed). Classification using random forest models and an unsupervised algorithm achieved an overall accuracy of over 0.85, compared to the accuracy of human experts for *E. elaterium* of over 0.74.

The study in [97] focused on developing a computer vision-based system for distinguishing potato plants from weeds in complex, high-occlusion environments during the post-emergence stage. A dataset of 1,950 RGB images from potato farms was collected, annotated at the pixel level, and made publicly available. Deep learning models, i.e., Mask RCNN and YOLOv8, were trained for weed detection, with YOLOv8 achieving a mean average precision of 83.4% and Mask RCNN reaching 79%. While YOLOv8 slightly outperformed Mask RCNN in overall mAP, Mask RCNN achieved higher precision, recall, and F1-score for the weed class, making it more effective for weed identification.

In [98], volunteer cotton weed plants growing amidst inter-seasonal and rotated crops, such as corn, become susceptible hosts for boll weevil pests upon reaching the pin-head square stage (5–6 leaf stage). Effective detection, localization, and targeted eradication or treatment of these weed plants are essential. This paper explored the application of machine/deep learning, specifically the YOLOv3 algorithm, to detect those weeds in corn fields using RGB images acquired by a UAV.

The importance of spatially explicit weed information for controlling infestations and minimizing corn yield losses is highlighted in [99]. UAV-based remote sensing offers an efficient approach to weed mapping, though thermal measurements (such as canopy temperature) have been underutilized. By integrating spectral, textural, structural, and canopy data, researchers identified optimal combinations for improved weed detection using machine learning algorithms. Results showed that incorporating canopy temperature and fusing textural, structural, and thermal features enhanced weed-mapping accuracy.

The research in [100] demonstrates how low-cost UAV platforms can effectively map giant smutgrass infestations in Florida bahiagrass pastures, enabling site-specific weed management and reducing herbicide use. RGB ortho-mosaics collected on two sampling dates (May and August) and at four different altitudes (50, 75, 100, and 120 m) were analyzed using spectral, texture, and combined approaches with both supervised and unsupervised classification methods. The best mapping results were achieved by integrating spectral and texture analyses with a supervised algorithm, yielding a correlation of 0.91 with ground truth data, although higher altitudes slightly reduced detection accuracy.

**Table 5**: The most common drone sprayers that have been or can be used in weed management.

| Drone application | Sprayer | Drone image |
|---|---|---|
| DJI, Agras T30 Hexacopter Orchard farm [101] | 30 L tank 16 nozzles 8 L/min | 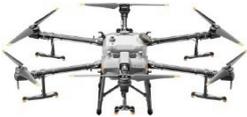 |
| DJI, Agras T40 Quadcopter Sugarcane fields [102] | 70 L tank 4 nozzles 12 L/min | 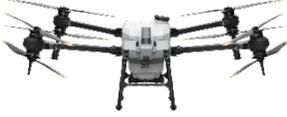 |
| Freeman, 2000 series Fixed wing Open-field farms [103] | 60 L tank 9 nozzles 4.4 L/min | 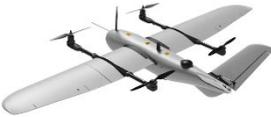 |
| XAG, P-series Quadcopter Cotton farms [104] | 15 L tank 4 nozzles 30 L/min | 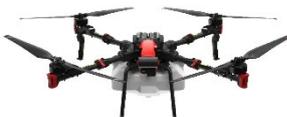 |
| XAG, V-series Bicopter Open-field farms [105] | 16 L tank 2 nozzles 10 L/min | 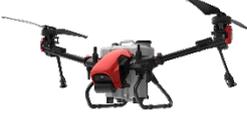 |
| Yamaha, Rmax Helicopter Pineapple farms [106] | 16 L tank 3 nozzles 8 L/min | 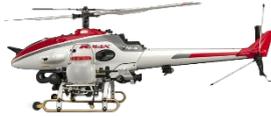 |

In addition to data collection, drones can be used for precision spraying. Table 5 provides a selection list of drones for weed spraying. The table highlights various drone models tailored for agricultural spraying, each designed to optimize efficiency based on specific farming needs. Multi-copter models (e.g., DJI, XAG, and Yamaha), suitable for smaller fields, feature different tank capacities and nozzle configurations to accommodate varying crop densities and in-flight maneuverability. Fixed-wing drones on the other hand (e.g., Freeman), are more suitable for open-field farms.

# IV. Data Processing

**Data**

Building on the data collection technologies and methods outlined in the previous section, large volumes of data can be gathered, requiring intelligent processing algorithms with advanced capabilities. These algorithms can be applied in a variety of management applications including monitoring vegetation health, identifying crop stress, detecting weeds and insect infestations, and enabling precise application of treatments such as herbicides, pesticides, or fungicides [107]. To effectively develop processing algorithms for these applications, large and diverse datasets, capturing variability across different domains and collected using the collection technologies discussed, are essential. This, in turn, gives rise to the big data challenge in agriculture.

*Big Data*

Big data refers to extremely large and/or diverse data types that are difficult to manage using traditional data processing tools. Agricultural data is especially getting big, due to the increasing use of technology like Internet of Things (IoT), drones, and satellites [108, 109]. Effectively handling heterogeneous agricultural data, such as environmental (temperature, humidity, rainfall), soil data (pH, moisture, nutrient levels), crop data (yield, health, growth stages, weed, pest, disease), and market data (prices, demand, supply), necessitates sophisticated data warehouses capable of storing, cleaning, standardizing, and integrating/fusing information from disparate sources [110]. Data storage and processing require scalable and cost-effective infrastructure, often leveraging cloud computing platforms. Hadoop [link] and other big data tools offer a promising solution to handle massive volumes of data generated in agriculture.

Cleaning data to remove noise and outliers, along with standardizing formats and protocols, is crucial for ensuring interoperability across diverse data sources. These processes enhance data quality and consistency, facilitating the aggregation and analysis of datasets from varied origins. As highlighted by Yu *et al.* [111], implementing reproducible data harmonization protocols (constructed from parameterizable primitive operations) enables transparent and scalable integration of heterogeneous weed mapping data. Such harmonized datasets support more effective comparisons, seasonal trend analyses, and accurate model training across different ecosystems/environments, aligning with the FAIR (Findable, Accessible, Interoperable, and Reusable) principles of data stewardship.

*Data Fusion*

Once cleaned and standardized, the data can be more easily fused to develop integrated systems that support precision agriculture, where timely and accurate information is critical. Data fusion techniques integrate information from multiple sources, including satellites, drones, sensor networks, and weather stations. By combining these diverse datasets, a more comprehensive understanding of field conditions is obtained, identifying key farming patterns, predicting risks, and enhancing the accuracy and reliability of agricultural decision-making [3].

Data fusion can be utilized for both imagery and non-imagery data types [112]. When it comes to imagery, various fusion methods are employed to integrate data from multiple sources, aiming to generate high-resolution images with enhanced spatial and spectral quality. A summary of the most used methods, their operational procedures, as well as their respective application in agricultural data handling, which have been or can be applied to weeds, is provided in Table 6. The table covers seven widely used methods: Brovey Transform, Intensity Hue Saturation (IHS), Principal Component Analysis (PCA), Wavelet Transform, Ehlers Fusion, and Gram-Schmidt (GS) Transform. These techniques aim to enhance spatial resolution while preserving spectral integrity by combining high-resolution panchromatic images with lower-resolution multispectral or hyperspectral images. Each method employs a distinct mathematical approach, ranging from spectral normalization and orthogonal transformations to frequency and component-based processing, to achieve effective fusion tailored to agricultural image analysis and precision weed detection.

In the context of agricultural monitoring, non-imagery data fusion incorporates a wide range of sources such as weather data, soil composition, water quality, pest and disease reports, historical yields, market prices, labor availability, etc. Multi-source integration of both imagery and non-imagery data enables a more holistic understanding of crop conditions, and environmental factors [99]. Weed detection particularly benefits from the utilization of fused data, offering effective differentiation between weeds and crops. For instance, Xu *et al*. [99] explored the use of data fusion for weed management by combining multiple types of spectral, textural, structural, and thermal measurements to improve weed mapping accuracy in corn fields. While thermal data (e.g., canopy temperature) had been underutilized, the research demonstrated that integrating it with other features significantly enhanced weed detection, boosting overall accuracy. The best performance was achieved by fusing textural, structural, and thermal features, with an machine learning model.

**Table 6**: Multi-source image data fusion techniques that have been or can be used in weed detection applications.

| Name | Application | Fusion process |
|---|---|---|
| Brovey Transform (BT) [113] | <ul><li>Merges high-res panchromatic and low-res multispectral images.</li><li>Enhances spatial detail while preserving spectral information.</li><li>Achieved by normalizing spectral bands and multiplying them with the panchromatic image.</li></ul> | Each band ($B_i$) of the multispectral image $S$ is transformed using a high-resolution PAN image as $$B'_i = \frac{B_i}{\sum B_n} \times PAN$$ The final fused image is reconstructed by combining these transformed bands $$S_{BT} = \langle B'_1, B'_2, \ldots \rangle$$ |
| Ehlers Fusion [114] | <ul><li>Implements a frequency-based fusion technique to RGB or multispectral images within the IHS color space.</li><li>Aims to preserve both spectral integrity and spatial resolution in hyperspectral and multispectral data.</li><li>Reduces spectral distortion compared to traditional fusion methods such as Brovey or standard IHS transformations.</li></ul> | Apply Fast Fourier Transform (F) to both the intensity and the high-resolution panchromatic images $$F_I = F(I)$$ $$F_{PAN} = F(PAN)$$ High-frequency components from the panchromatic image are selectively added to the intensity component (by adaptive filtering) to form $F'_I$. Finally, apply Inverse FFT to return intensity back to the original space $$I' = F^{-1}(F'_I)$$ $$S_{Ehler} = \langle I', H, S \rangle$$ |
| Gram-Schmidt (GS) [115] | <ul><li>Enhances the spatial resolution of multispectral images using advanced image fusion techniques.</li><li>Maintains spectral integrity during the fusion process to ensure accurate color representation.</li><li>Uses these techniques in remote sensing and satellite image processing applications.</li></ul> | Each multispectral band ($B_i$) is transformed into an orthogonal basis using the GS process, which ensures that each new component ($B'_i$) is uncorrelated with the previous ones. The first transformed component ($B'_1$) is replaced with a high-resolution panchromatic image. The fused image is then reconstructed by the inverse GS transform $$S_{GS} = GS^{-1}(PAN, B'_2, \ldots)$$ |
| IHS [116] | <ul><li>Enhances the spatial resolution of RGB images to improve visual detail.</li><li>Preserves the natural color of the images during the enhancement process.</li><li>Acknowledges the potential for spectral distortion as a limitation of the method.</li></ul> | The process includes (1) converting RGB to IHS, (2) replacing the Intensity component, $I$, with a high-resolution panchromatic image, and (3) converting the new HIS image back to RGB. |
| Principal Component Analysis (PCA) [99] | <ul><li>Converts correlated multispectral image bands into a smaller set of uncorrelated components using PCA.</li><li>Reduces data dimensionality by extracting principal components that capture the most significant variance.</li><li>Utilizes the first principal component to carry and enhance spatial details of the image.</li><li>Preserves spectral information while improving spatial resolution through component substitution.</li></ul> | For an input multispectral image $S$ with the matrix of eigenvectors $V$, the PCA is $$PCA = V^T S$$ We then replace the first $PCA$ component with a high-resolution panchromatic image to form $PCA'$. Finally, we transform back the $PCA'$ into the original spectral space. |
| Wavelet Transform [117] | <ul><li>Combines high-resolution panchromatic images with low-resolution multispectral images using a powerful fusion technique.</li><li>Avoids simple arithmetic fusion methods that may compromise image quality.</li><li>Preserves both spatial and spectral details effectively through wavelet-based fusion.</li></ul> | Both the panchromatic and multispectral images are decomposed into low-frequency (approx.) and high-frequency (detail) wavelet components. The low-freq. of the panchromatic image will be added to the high-freq. multispectral coefficients. The fused image will then be reconstructed by an inverse transformation. |

Another noteworthy work is conducted by Xia *et al.* [118], where they introduced a novel approach to weed resistance management by developing a comprehensive resistance score and using multimodal data sources, i.e., spectral, structural, and textural, to map herbicide-resistant weeds. By employing deep learning and various fusion strategies, especially late deep fusion models, the researchers enhanced resistance assessment accuracy. The hyperspectral data proved most informative individually, but combining all modalities coupled with deep learning, significantly improved regression performance across different weed densities.

*Citizen Science*

Citizen science in agriculture involves the active participation of non-specialists, such as farmers, in scientific research processes. This approach leverages the collective power of individuals to gather data, conduct experiments, and contribute to agricultural innovation. By engaging citizens, researchers can access vast amounts of localized data that would otherwise be difficult or expensive to collect. In agriculture, citizen science has been particularly valuable for on-farm testing of crop varieties, monitoring environmental conditions, and assessing pest and weed infestations.

Table 7: Publicly available annotated weed image datasets.

| Dataset name, publication year | Modality | Image Count | Annotation method |
|---|---|---|---|
| **Gathered by handheld devices** | | | |
| Carrot-weed, 2018 | RGB camera | 39 | Segm. [link] |
| Leaf counting, 2018 | RGB camera | 9,372 | Count [link] |
| Early-crop-weed, 2019 | RGB camera | 508 | Class. [link] |
| DeepWeeds, 2019 | RGB camera | 17,509 | Class. [link] |
| **Gathered by vehicles and robots** | | | |
| Crop/weed field image dataset, 2015 | MS sensor | 1,500 | Segm. [link] |
| Sugar beets, 2016 | NIR and RGB camera | 25,429 | Segm. [link] |
| Crop vs weed discrimination, 2019 | NIR and RGB camera | 40 | Segm. [link] |
| Ladybird cobbitty brassica, 2019 | Thermal, HS, RGB, weather, and soil data | 2,245 | Class. [link] |
| Open plant phenotyping of weeds, 2020 | RGB camera | 7,590 | Det. [link] |
| The Rosario dataset, 2022 | Stereo images and GPS positional data | 15 per second | Det. [link] |
| Phenotyping in Weed Identification, 2024 | RGB camera | 28,000 | Class. [link] |
| Weed-crop, 2025 | RGB camera | 3,020 | Class. [link] |
| **Gathered by drones** | | | |
| Grass and broadleaf weeds, 2017 | RGB camera | 400 | Segm. [link] |
| WeedNet, 2018 | NIR sensor | 465 | Segm. [link] |
| Columbia invasive species, 2018 | RGB images and GPS positional data | N/A | Det. [link] |
| *Cynodon dactylon* in vineyard, 2019 | Photomosaic of RGB images | N/A | Segm. [link] |
| Weed detection projects, 2022 | RGB camera | 4,201 | Det. [link] |
| SeSame, weed aerial dataset, 2023 | RGB and NDVI camera | 1,920 | Det. [link] |
| Tobacco Dataset for crop/weed classification, 2023 | RGB camera | 1,600 | Segm. [link] |
| Sandplain lupin weeds, 2023 | Photomosaic of RGB images | 578 | Det. [link] |
| Broad-leaved pepper weed, 2024 | MS sensor | 26,763 | Segm. [link] |
| DroneWeed, 2024 | RGB camera | 31,002 | Det. [link] |
| **Gathered from various sources** | | | |
| CottonWeedID15, 2023 | RGB camera | 584 | Segm. [link] |

Segm: Segmentation, Det: Detection, Class: Classification
NIR: Near-infrared, MS: Multispectral, HS: Hyperspectral, N/A: Not Available

For instance, initiatives such as ClimMob [119] have created software to simplify experimental design and data collection, allowing farmers to engage in large-scale trials that support agricultural practices like on-farm testing and experimental citizen science. The proliferation of smartphone technology has further enhanced this approach, allowing farmers to easily document and share observations, such as weed presence and crop health, in real-time [120].

Weed mapping is a critical application of citizen science in agriculture, as it provides spatial and temporal insights into weed distribution and density. Traditional weed mapping methods are labor-intensive and often limited in scope. However, citizen science can scale up data collection by involving farmers and the public in recording weed species and their locations across large areas. In this regard, geostatistical techniques combined with GPS-enabled devices [121] can been used to map weed populations in non-tillage systems.

The potential of citizen science for weed mapping extends beyond data collection to fostering collaboration between farmers, researchers, and policymakers. By involving farmers in the research process, citizen science projects can generate locally relevant solutions that are more likely to be adopted. Moreover, the data collected can inform sustainable weed management strategies, such as ecological redesign of cropping systems and the use of microbial nitrogen immobilization to suppress weed growth [122].

*Data Annotation*

A fundamental objective shared across computer vision-based precision agriculture tasks is the accurate detection of specific objects of interest, e.g., weeds, crops, or fruits, while distinguishing them from the surrounding environment. Achieving this not only depends on well-designed model architecture and reliable hardware implementations but also requires robust supervised or semi-supervised data. This typically involves training machine learning models on carefully annotated images to enable accurate and consistent identification [123].

Image annotation is the process of labeling sufficiently large image sets with meaningful semantic information, which is crucial for training AI models. Creating large-scale annotated datasets is a challenging and resource-intensive task. It involves significant effort and cost for image collection, categorization, and annotation, as well as, in some cases, physicochemical analysis of crops [123]. One practical solution to these challenges is data sharing, which holds exciting potential for accelerating scientific advancements. Publicly available datasets not only reduce the time and cost associated with dataset preparation but also facilitate the benchmarking of image analysis and machine learning algorithms across different research groups.

The computer vision community has benefited from the availability of public annotated image datasets, which have driven major advances in object detection, segmentation, and the development of innovative model architectures. While there are several plant-specific image datasets available, many are not directly applicable to weed mapping. A collection of relevant annotated weed datasets is summarized in Table 7. This table presents a diverse collection of agricultural image datasets focused on weed monitoring, categorized by the method of data acquisition, i.e., handheld devices, ground vehicles/robots, drones, and mixed sources. These datasets span from 2015 to 2025 and cover a range of modalities, including RGB, NIR, spectral, and thermal imaging. Vehicle- and robot-acquired datasets are generally larger and more multimodal. Although some of these datasets additionally include GPS coordinates for weed localization, such geospatial metadata is not strictly necessary for designing and training accurate machine learning models.

Although several open-source weed datasets exist, there remains a significant gap in the availability of comprehensive, high-quality foundational datasets tailored to specific crops. While foundational AI models have achieved remarkable success in other domains, replicating this progress in agriculture requires large, diverse, and crop-specific datasets [124], for example, datasets focused on broadleaf weeds. The recent development of WeedNet [125] demonstrates promising progress towards global-scale weed species identification using a foundational model approach. However, despite its achievements, WeedNet also highlights the ongoing need for targeted, regionally adapted datasets and models that capture the nuances of specific cropping systems and agroecological contexts.

If ready-to-use datasets are not available for a new application, then new image datasets must be gathered, and image annotation tools need to be employed. Traditional annotation methods are usually manual, which make them time-consuming and labor-intensive, hence, impractical for large-scale agricultural datasets. Modern image annotation tools and techniques, on the other hand, enable significant advancements in precision applications by facilitating dataset creation. These tools can automate or semi-automate the annotation process, improving efficiency and accuracy. They often incorporate interactive elements, allowing users to refine annotations and correct errors, thereby enhancing the quality of the training data [126].

Semi-supervised learning techniques leverage both labeled and unlabeled data to improve active annotation learning model's performance, particularly when labeled data is scarce [127]. Transfer learning approaches utilize pre-trained models on large, general-purpose datasets and fine-tune them for specific agricultural tasks, accelerating the training process and improving accuracy. Furthermore, novel techniques like propagating labels from semantic neighborhoods can address issues such as class imbalance and incomplete labeling, common problems in agricultural datasets [128]. These techniques are combined with modern AI-assisted annotation tools, such as the Computer Vision Annotation Tool (CVAT) [link], Roboflow Annotate [link], and MakeSense [link], to provide efficient and accurate annotations for agricultural applications.

In contrast to all these closed vocabulary techniques, open-vocabulary semantic segmentation, enhanced by Large Language Models (LLMs), represents a significant advancement in few-shot segmentation of weeds/crops [129]. Close-vocabulary weed annotation techniques rely on a limited set of object classes, constraining their ability to identify new or unseen weed species.

Open-vocabulary approaches, however, leverage the semantic knowledge embedded in LLMs to recognize a broader range of plant species without requiring extensive retraining. This is achieved by aligning visual features with rich semantic features learned from vast amounts of text and image data [129]. For example, models like CLIP [link], which are pre-trained on large-scale vision-language datasets, can be adapted to segment images based on textual descriptions of weed characteristics, even if those specific species were not present in the original training data [130]. This capability is crucial in agricultural settings where weed populations are diverse and constantly evolving. The integration of LLMs allows for more flexible and adaptable weed mapping systems that can respond to new challenges and changing environmental conditions.

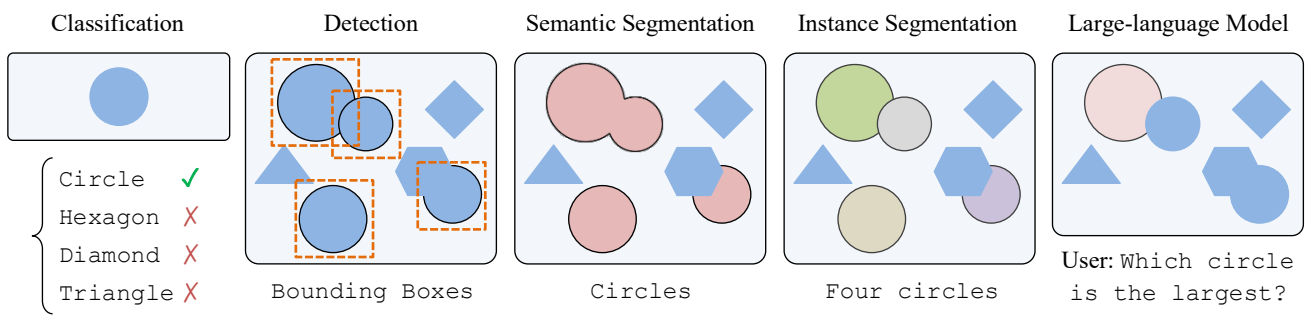

**Figure 5**: Conceptual comparison between the most common machine learning approaches in data processing.

Furthermore, Few-shot segmentation, a technique designed to perform image segmentation with minimal training examples, is particularly useful in weed mapping due to the excessive cost and effort associated with acquiring labeled data. By combining LLMs with few-shot learning techniques [130], researchers can develop robust weed mapping systems that require only a handful of annotated images to accurately segment different weed species.

**Machine/Deep Learning**

Machine Learning (ML) and Deep Learning (DL) are foundational to the advancement of modern weed mapping technologies in agriculture. These computational methods have significantly outperformed traditional approaches in terms of detection accuracy, cost efficiency, and implementation adaptability. By leveraging intelligent algorithms, ML and DL facilitate various tasks such as weed identification, spatial mapping, resource optimization, and automated treatment strategies [131].

This section explores the major applications of ML in weed mapping, organized into four key areas: classification, detection, segmentation, and LLMs. These key application areas are illustrated in Figure 5. Classification models can identify the presence or absence of weeds in an image [132], but lack precise spatial information. Object detection models locate weeds by drawing bounding boxes around them, providing spatial coordinates but limited pixel-level detail [80]. Semantic and instance segmentation models classify each pixel in an image as either weed or crop, generating detailed weed maps and facilitating precise herbicide application [132]. While the application of LLMs in weed mapping is nascent, their potential lies in integrating contextual information with image data to answer end-users' queries. This can be particularly important in analyzing farmer's field notes and improving weed prediction and management [130].

*Classification*

As stated before, weed classification involves categorizing different plant species, particularly distinguishing between weeds and crops, from images or sensor-derived data. This is essential for species-specific control and effective weed management strategies. Traditional Machine Learning: Algorithms like K-Nearest Neighbors (KNN), Random Forest, and Decision Trees remain effective for smaller datasets or environments with limited computational capabilities. These methods require manual feature extraction, such as color, shape, and texture descriptors, and are still viable for initial feasibility studies or resource-limited settings [41].

Modern DL architectures such as ResNet, EfficientNet, and Vision Transformers (ViTs) have demonstrated exceptional accuracy in plant classification tasks. These models automatically learn complex visual features and patterns from large agricultural image datasets, offering improved performance over handcrafted feature methods. Lightweight Convolutional Neural Networks (CNNs) are also widely used [133], especially for scenarios with complex backgrounds or constrained hardware resources [8].

*Detection*

Weed detection focuses on locating the presence and position of weeds within an image or field. A variety of architectures are employed, with You Only Look Once (YOLO) variants, including YOLOv3 to YOLOv10, being particularly popular due to their speed and efficiency in real-time applications. For instance, when detecting volunteer cotton weed plants in corn fields, YOLOv3 achieved an average detection accuracy exceeding 80%, with an F1-score of 78.5% [98]. A study in 2025 compared YOLOv5 and YOLOv8 for weed detection in cotton farming, highlighting their effectiveness in identifying weeds that compete with cotton crops [134]. Furthermore, research explores modifications and enhancements to the YOLO architecture, such as the PMDNet model built upon YOLOv5, designed for efficient weed detection in wheat fields [135].

Beyond the YOLO family, other deep learning architectures are also being explored for weed detection. Region-based Convolutional Neural Networks (RCNNs) have been applied to detect and classify weeds in potato field, demonstrating the potential of these models in specific agricultural contexts [97]. ViTs are also being considered as effective DL architectures, where these attention-based models are implemented as intelligent weed control system in natural corn fields [136].

*Segmentation*

Weed segmentation involves partitioning an image into distinct regions/pixels corresponding to crops and weeds. This provides a more detailed understanding of weed distribution and density compared to detection alone, allowing for precise herbicide application, reducing overall chemical usage, minimizing environmental impact, and increasing/estimating yield [137]. Segmentation models are widely integrated into weed management robots and UAVs for automated weed detection and targeted herbicide application [72].

Recent research has focused on developing and improving segmentation models to address the specific challenges of weed detection, such as the similarity in spectral features between crops and weeds, variations in weed growth stages, and complex field environments. CNNs, particularly EfficientNet-based models and encoder-decoder architectures like U-Net, DeepLabV3, and PSPNet, are widely used for this advanced computer vision purpose [138].

Farmers usually plant a specific type of crop in their farms. Some studies use this opportunity and focus on segmenting the crop(s), and then classifying the remaining green objects as weeds to reduce model complexity [139]. Researchers are also exploring attention mechanisms and feature fusion techniques to improve segmentation accuracy and robustness in challenging field conditions [140].

Data augmentation techniques to increase the size and diversity of training datasets [141], synthetic data generation (i.e., creating realistic training samples) by pasting segmented plant patches onto soil backgrounds to address the scarcity of labeled data [142], and transfer-learning approaches to leverage knowledge from existing datasets and improve model performance in new environments or with different crop types [143] are among the other weed segmentation improvement solutions.

*Large Language Models*

LLMs are increasingly being explored for their potential to revolutionize various aspects of the agricultural sector, including weed management. They offer a promising avenue for automating and enhancing annotation delays, especially without human expert involvement, leading to more efficient and targeted weed control strategies [144]. These models can integrate image features from DL models with textual contexts from natural language processing models to offer a unified query-able neural network.

LLMs are also being used to enhance named entity recognition for agricultural commodity monitoring, which indirectly impacts weed management [145]. Indirect weed detection has previously described as detecting crops first and then naming other green objects as weeds. Similarly, by pretraining transformer-based language models with food-related textual data, semantic matching between food descriptions and crop images can be established, offering insights into potential weed objects [146]. This approach can be expanded to identify and classify weeds based on textual descriptions and associated data.

The combination of Reinforcement Learning (RL) and LLMs represents a novel approach with transformative potential in the agricultural sector, offering adaptive strategies. In research conducted by Chen *et al*. [147], the study emphasizes the importance of efficient and sustainable crop production management, aiming to minimize environmental impacts through RL-LLM integration. Traditional methods struggle to adapt to the evolving dynamics influenced by climate change, soil variability, and market conditions, whereas RL-LLM integration has enhanced crop management decision support systems by optimizing decision-making through data-driven approaches. Despite considerable progress, challenges related to real-world deployment complexities remain.

**Edge Processing**

Edge processing, defined as the deployment of ML and DL models on local devices rather than relying on cloud-based computing, holds significant promise in agricultural applications such as weed mapping. Traditional approaches often require extensive computational resources and suffer from latency issues when processing substantial amounts of data from remote locations. By contrast, edge processing allows for real-time analysis directly at the source, which is crucial for off-grid, time-sensitive, and/or continuous detection tasks in field vehicles/robots [148]. This capability is especially advantageous in environments where real-time monitoring and immediate response are necessary.

In weed mapping, edge processing offers several advantages. Firstly, it enables rapid identification and classification of weeds directly from captured images or sensor data. By deploying detection models locally on edge devices like vehicle-mounted cameras or drones, farmers can quickly assess the extent of weed infestations without relying on external networks, thus reducing dependency on internet connectivity and cloud services [149]. Additionally, real-time monitoring enabled by edge devices ensures that weed management strategies can be adjusted in-time based on evolving field conditions.

Edge processing has certain limitations, such as lower throughput, limited memory that restricts model complexity, and constrained energy availability. Nonetheless, it plays a crucial role in enabling faster and more precise weed control strategies. Once weeds are identified on edge, immediate management action can be taken. This approach not only enhances operational efficiency but also supports sustainable farming practices by reducing chemical usage and preserving soil health [149].

# V. Weed Mapping

**Spatiotemporal Patterns**

Understanding the spatial and temporal distribution of weed species within agricultural and environmental systems presents a complex challenge due to the inherent heterogeneity of agroecosystems. Variability in weed distribution arises from both regional and local factors. At the regional scale, differences in climate, field management histories, landscape structure, and soil composition contribute to weed diversity. Locally, factors such as farmer expertise, cultural practices, soil characteristics, topography, and microclimatic conditions significantly affect weed emergence and distribution. Temporal dynamics of weed distribution are also essential for optimizing long-term management strategies. Weed patches keep changing spatially over time, including the annual changes in patch boundaries, and instabilities in their distributions [150].

Weed communities are shaped by the ecological requirements of species, such as growth form, phenological development, and sunlight requests. These traits, coupled with agricultural management practices, lead to noticeable spatial clustering of weeds within fields [151]. Numerous studies have documented that many weed species exhibit aggregated spatial distributions, which means they form patches rather than spreading uniformly [152]. Despite this, herbicides are often broadcast across fields, leading to overuse and environmental harm. Based on a study by Blank *et al.* [150], 86% of weed species exhibited patchy distributions. Aggregated patterns were dominant in key weed genera such as *Avena*. In contrast, some other genres like *Chenopodium* were found to be randomly distributed.

Seeds' weight, morphology, aerodynamic, and parent's height are other factors that influence the spatial distribution of weeds. Vegetative reproduction, as well as granivores further reinforce aggregation. Overall, most weed seeds fall near the parent plant, especially those without wind or water-assisted dispersal mechanisms, e.g., *Ecballium elaterium*, which further emphasizes the benefit of weed mapping before any weed control application [153]. Conversely, wind-dispersed seeds, e.g., *Taraxacum officinale*, may produce more randomized distributions.

Temporal weed mapping is as important as its spatial patterns. However, 63% of studies spanned only one to two years, making them insufficient for assessing long-term temporal weed trends. Only 6% extended beyond five years. Species with wind-dispersed seeds or low population density tend to show less temporal stability. Understanding temporal trends in time-based weed mapping allows for strategic pre-emergence and post-emergence herbicide applications based on historical data. For persistent weed patches, farmers can timely localize their pre-emergent herbicides, thus optimizing product efficiency. The exact timing of post-emergence treatments also depends on the weed emergence pattern to avoid inefficiencies and off-target effects [150].

**Farm Management Effects**

Farm management practices significantly influence weed distribution by altering soil conditions, crop rotation patterns, and disturbance regimes. In this regard, cropping systems (i.e., crop type and its associated management practices) heavily influence the weed population and distribution dynamics. Crop canopy architecture, growth vigor, and competitive traits affect weed suppression [154]. For example, maize creates dense shade that limits weed growth, while onion with slow growth and weak canopy cover is a poor competitor. Mechanical cropping operations, like harvesting, also affect seed dispersal and subsequent weed distribution. For example, combine harvesters can spread seeds along the direction of travel, contributing to elongated weed patches [155].

Blank *et al.* [150] report that 97% of weed mapping studies focused on broadacre crops, with only 1.5% each in orchards and vineyards. Corn and wheat were the most frequently studied crops, comprising 27% and 23% of studies, respectively. Aggregated weed patterns were most common across major cereals, including maize, wheat, soybean, and barley. Crop competition traits, i.e., early canopy closure, tillering, and root expansion, affect the composition of weed communities. For example, dense crops may favor climbing species like convolvulus, while open-canopy crops benefit rosette-forming weeds. As a result, the need for frequent weed mapping is greater for variable farming practices, as well as species with unstable distributions. For example,

- In rotating crops, the variability in field conditions leads to shifts in weed patch dynamics over time.

- In orchards, where UAVs cannot look under the treetops, understanding patch stability may be even more critical for effective weed management.

Another important farm management practice is the use of herbicides, including their types, dosages, and methods of application. Mapping herbicide usage and herbicide-resistant weeds is just as important as mapping the weeds themselves. Herbicides account for a huge portion of global weed control strategies, but the overuse of specific modes of action has led to widespread herbicide resistance in many weed species. Monitoring and mapping the occurrence of herbicide-resistant weeds are essential for timely detection and effective resistance management.

Table 8: Comparing the software tools commonly used in agricultural science and weed mapping.

| Tool | Description | Cost | Features | Use Cases |
|---|---|---|---|---|
| Agisoft Metashape | Photogrammetry software for 3D point clouds, orthophotos, and terrain models | Moderate | 3D modeling, DSMs, vegetation structure analysis | UAV terrain modeling, weed/disease distribution [74] |
| ArcGIS Pro | Industry-standard GIS platform for advanced 2D/3D mapping and spatial analysis | Expensive | Advanced 2D/3D, spatial analysis, mobile/GPS integration | High-end research, enterprise-level Ag data, weed Treatment [156] |
| DroneDeploy | Cloud-based drone mapping platform with AI analysis and Ag modules | Moderate | NDVI, plant health, automatic report generation | Commercial Ag, UAV-based weed maps [81] |
| ENVI | Remote sensing software to process hyperspectral and multispectral images | Expensive | Spectral analysis, NDVI, land cover classification | High-end research, spectral weed detection [6] |
| ERDAS Imagine | Professional image processing tool for raster and satellite data analysis | Expensive | Raster modeling, remote sensing, 3D terrain analysis | Soil/vegetation analysis, GIS labs, weed segmentation [157] |
| Field Maps / Survey123 | Mobile GIS apps for collecting georeferenced field data | Free to moderate | Offline mapping with GPS, data collection forms | Ground-truthing, weed survey/management [158] |
| Google Earth Engine | Cloud-based large satellite data analysis using code | Free for research | Massive data library, time series analysis | Remote sensing, regional weed/crop monitoring [83] |
| Pix4D | Drone image processing software for creating maps and 3D models | Moderate to High | Orthomosaics, NDVI, 3D modeling, photogrammetry | Field scouting, UAV-based weed maps [81] |
| QGIS | Open-source software for spatial analysis and mapping | Free | Plugin support, raster and vector analysis, basic 3D | Academic research, weed and disease mapping [74] |
| SST Summit / SMS Advanced | Precision ag software for analyzing field data, generating zones, and prescriptions | Moderate to High | Yield maps, variable rate, field analysis tools | Precision farming, crop and weed management [159] |
| Trimble Ag / Farm Works | Integrated farm management software with GPS and variable rate technology | High | GPS, soil/plant data, input prescriptions | Farm decision support, weed detection [160] |

Tools such as geo-referenced databases and interactive web-based platforms enable researchers, advisors, and policymakers to visualize the spread of resistant populations, facilitating more targeted and sustainable weed control strategies [161]. For example, Weedscout 2.0 has been developed to track herbicide-resistant *Alopecurus* species across parts of Europe, while the iMAR system in Italy allows for continuous updates and visualization of resistance data in *Echinochloa* species [162].

Global efforts to maintain an accurate database of herbicide-resistant weed cases are led by the International Herbicide-Resistant Weed Database [link], offering detailed maps based on herbicide mode of action. Tools like this are not only useful for farmers and researchers but also for policymakers designing integrated weed management frameworks, ensuring that herbicide application remains effective and sustainable.

As can be seen, enhancing farm management, particularly in weed control, requires software-based spatiotemporal data visualization on weed distribution, soil conditions, and crop health. Satellite, drone, and field observation data can be visualized by GIS maps and tools, providing a strong decision-support basis in agriculture. These tools will be studied in detail in the next section.

**Common Maps and Tools**

2D and 3D thematic maps are essential tools in weed mapping applications, for visualizing spatial data, analyzing field conditions, and supporting precision management. A breakdown of the most common types of 2D maps used in this context include: Choropleth Maps to display variations of a variable (e.g., weed density or zoned statistics) using color gradients; Dot Density Maps to represent frequency of features (e.g., weed populations or distribution patterns) with dots; Isoline/Contour Maps to connect points of equal value (e.g., farm/land topography or soil parameters levels) by lines; Raster Maps of grid cells (e.g., weed vigor or vegetation indices) where each cell holds a value; Symbol Maps with proportional symbols and their sizes (e.g., weed biomass or herbicide intensity) according to data magnitude; and Heat Maps to represent data density or intensity (e.g., weed infestation zones or hotspots) with color gradients.

Similarly, the most common 3D Maps in weed mapping applications include: Digital Elevation Models to visualize elevation and topography (in weed intensity studies, water flow calculations, irrigation planning, and soil conservation acts); Point Cloud Maps from Light Detection and Ranging (LiDAR) or UAVs to visualize field surfaces (in weed structure analyzes, weed biomass estimation, and canopy analysis); 3D Vegetation Index Maps to combine remote sensing data with height models to give a volumetric perspective (in weed-crop competition assessment and plant health monitoring); and 3D Time Series Maps to show changes over time with height and intensity layers (in weed or crop growth evaluation and temporal weed dynamics/development).

A comparison between the key software packages in agricultural science and weed mapping is presented in Table 8. This table summarizes the most common tools based on their cost, ease of use, key features, and application scenarios. This comparison is helpful in choosing the right tool depending on project's needs, whether it is academic research, commercial farm management, or field surveying.

# VI. Future Directions

Although modern technologies for weed mapping have advanced significantly, many barriers are left unaddressed, preventing these advancements from being used in real-world applications. Key challenges include the lack of practical and cohesive data and outdated hard and soft technologies. This underscores the need for balanced and unbiased data collection, and modern deep learning analysis, and more intuitive weed mapping techniques to meet diverse agricultural demands. This section explores some of these major challenges in depth, highlighting key opportunities for advancing data-driven solutions to support the evolving needs of precision weed management.

**Data Acquisition**

*Environmental Diversity*

Comprehensive annotated datasets that encompass various weed growth stages and environmental conditions will enhance ML model robustness and generalizability across diverse agricultural settings. Future studies should incorporate multi-regional trials that consider environmental variables such as climate, vegetation types, and seasonal shifts [91]. Trials and data collection conducted across different regions and at varied times of day will increase the robustness and transferability of weed detection models. Besides, there is a pressing need for long-term and multi-season data collection to better understand temporal weed distribution and to evaluate management strategies over time.

*Early-stage Data*

Accurate detection of weeds and plant diseases in their early development stages is still limited. New research should develop technologies capable of collecting data and identifying weeds before outbreak being visible, potentially using machine learning-enhanced remote sensing methods [105]. This also counts for sparse weed densities, especially at lower thresholds, to prevent misclassification and enhance weed resistance evaluation accuracy. Accurate detection of weeds in their early development stage would also translate into more effective weed control as younger weeds are easier to control with herbicides at lower rates.

*Remote Sensing Constraints*

UAVs often lack the space and frequency resolution sensors necessary for precise weed identification. Future advancements are expected to focus on wide-band or multi-band, as well as high-resolution hyperspectral and multispectral sensors to enhance the precision of weed identification [96]. Compared to RGB cameras, spectral sensors provide richer information, enabling more accurate discrimination of plant species.

In addition to sensor integration, optimizing the power consumption of UAV embedded systems is crucial for developing low-cost, long-endurance drones suitable for high-range agricultural applications. Sensors and onboard processing units can be energy-intensive, limiting flight times and operational efficiency. By optimizing both hardware and software components for energy efficiency, future UAVs can achieve longer flight durations, covering larger areas.

*Internet of Agricultural Things*

Advancing the use of IoT networks, including Bluetooth Low Energy, Radio-Frequency Identification (RFID), and IP-based sensors will allow better data collection, tracking, and monitoring of weeds and associated biosecurity threats across the agricultural supply chain. Towards this end, solving interoperability issues between devices, platforms, and datasets is critical [4]. Open-source standards and platform-agnostic data formats will facilitate smoother integration and decision-making across the agricultural ecosystem.

*Citizen Science*

An emerging and highly scalable approach to addressing the challenges of weed control data collection is the integration of citizen science with smartphone-based imaging. With over five billion unique mobile subscribers worldwide, engaging local communities in image data collection offers a cost-effective and logistically feasible alternative to conventional methods [120]. However, image quality and consistency remain critical challenges.

**Data Processing**

*Data Fusion Techniques*

Especially in the realm of spectral imaging, the future of weed detection is centered around the fusion of multispectral and hyperspectral data with deep learning methodologies. The utilization of vegetation indices, such as Normalized Difference Vegetation Index (NDVI) and Green NDVI (GNDVI), derived from spectral bands, provides valuable information on plant health and stress levels, aiding in the discrimination between crops and weeds. In the meantime, the fusion of multi-source data (e.g., UAV, satellite, IoT) offers promise for high-resolution, real-time weed mapping [9]. Future research should prioritize data fusion models that leverage deep learning to integrate multi-source spatiotemporal data seamlessly.

*Image Annotation & Segmentation*

Robust annotation tools that can manage occlusions, lighting variations, crop diversity, and the complex morphology of weeds are necessary. Advancements in automated annotation methods, such as semi-supervised learning frameworks utilizing adversarial strategies, have shown promise in reducing the manual effort required for pixel-level annotations [163]. Additionally, the integration of multi-sensor segmentation techniques, combining data from RGB, multispectral, and hyperspectral sensors, can enhance the accuracy of weed identification by leveraging the strengths of each modality. AI-assisted annotation platforms, like those employing superpixel algorithms, offer interactive and efficient means to annotate complex plant structures, thereby accelerating the creation of high-quality annotated datasets.

*Generative AI*

Generative AI offers a solution to the challenge of data scarcity in weed mapping by enabling the creation of synthetic datasets that mimic real-world conditions. Techniques such as diffusion models and generative adversarial networks can generate high-fidelity images of various weed species under different environmental conditions, enhancing the robustness of detection models. These synthetic datasets can be used to train deep learning models, improving their performance in real-world scenarios where annotated data is limited. Additionally, combining synthetic data with real-world data through domain adaptation techniques can further enhance model generalization [141]. Nonetheless, challenges remain in ensuring the realism of synthetic data and its alignment with actual field conditions, necessitating ongoing research to refine these methods.

*Advanced Models on the Edge*

The deployment of efficient and lightweight models, such as YOLO and Region-Fusion Detection Transformer (RF-DETR), on edge devices like drones and autonomous ground vehicles is anticipated to facilitate on-the-fly weed identification and mapping. This real-time capability is crucial for implementing precision agriculture practices, enabling timely and targeted weed management interventions [148]. The incorporation of ensemble learning techniques is also expected to improve detection accuracy by combining predictions from multiple models, thereby mitigating the limitations of individual models in complex field scenarios [138]. Furthermore, the integration of temporal data through time-series analysis is expected to capture the phenological changes of vegetation, enhancing the detection of weed emergence patterns over time.

*Vision Language Models*

The integration of VLMs into weed mapping presents promising avenues for enhancing annotation efficiency and detection accuracy. VLMs can assist in automating the weed annotation process by interpreting complex weed imagery, thereby reducing the reliance on manual labelling. This capability is particularly beneficial in scenarios involving occlusions and diverse crop types. Moreover, VLMs can be fine-tuned to understand the nuances of different weed species, enabling more precise identification and classification [147]. However, to ensure reliability and affordability, further research is needed to optimize these models for agricultural applications, considering factors such as computational resources and the need for explainable AI to gain trust among end-users.

**Mapping and Interpretation**

*Spatiotemporal Distribution Modelling*

Future works need to expand spatial pattern analysis across diverse geographies and multi-year timelines. Current models often lack the capacity to capture the dynamic nature of weed populations over space-time, limiting their effectiveness in long-term management strategies. This modelling is essential for understanding the persistence and evolution of weed populations in varying agricultural landscapes [150]. Moreover, future research should focus on improving seed dispersal modelling, including natural [153] and equipment-driven [155] mechanisms for better understanding/interpreting spatial distribution of weeds. Additionally, the effect of other factors such as soil type, moisture levels, and topography, on weed distribution, establishment, and proliferation need to be studied [152]. Integrating these variables into spatiotemporal models can provide a more comprehensive understanding of weed dynamics, leading to more effective and site-specific management strategies.

*Real-Time Decision Support*

Future advancements in real-time decision support systems for weed control can integrate advanced detection techniques and weed density and distribution models to facilitate site-specific management strategies. By leveraging technologies such as UAVs, IoT, and ML models, these systems can provide farmers with timely, actionable insights tailored to their specific field conditions [105]. These decision support systems should incorporate user-friendly interfaces to ensure that farmers, regardless of their technical expertise, can interpret and act upon the data effectively. Moreover, the integration of predictive analytics allows for proactive weed management, optimizing resource allocation and minimizing environmental impact.

*Global Biosecurity Governance*

Weed mapping as a biosecurity measure requires a multilateral governance approach. Establishing international conventions [2] and promoting open-data ecosystems [120] will foster collaboration and accelerate response to invasive threats. Future research should focus on making global weed mapping information affordable and accessible to small-scale farmers. This includes the development of user-friendly and mobile platforms, along with farmer-centric training programs.

# VII. Conclusion

This review systematically explored the landscape of weed mapping by analyzing the latest advancements in data acquisition, processing, and mapping techniques. We identified the major sensing platforms, ranging from handheld and vehicle-mounted devices to UAVs and satellites, and evaluated their integration with RGB, spectral, NIR, thermal, and terahertz imaging technologies. In the data processing domain, we reviewed deep learning-based approaches for data annotation, weed classification, detection, and segmentation, as well as the emerging use of edge computing and large language models for real-time, in-field processing. By focusing on spatial and temporal weed dynamics, as well as the influence of farm management practices, this review also shed light on the essential role of GIS-based mapping tools in supporting informed and targeted weed control decisions.

Importantly, this work fills a critical gap in the literature by being the first systematic review dedicated solely to weed mapping, following the PRISMA methodology to ensure methodological strength and transparency. The findings serve as a comprehensive knowledge base for scientists, Agri-tech developers, and decision-makers, helping them understand current capabilities, limitations, and opportunities for innovation. The insights presented herein not only guide future research in the design of smarter, data-driven weed management systems, but also support the broader goal of sustainable agriculture through reduced chemical usage and enhanced crop and environment health. As such, this review is positioned to influence both scientific inquiry and practical implementation in the evolving landscape of precision weed management.

# Acknowledgements

This research is supported by the Australian Research Council Industrial Transformation Research Program (ITRP) through the Training Centre in Plant Biosecurity (IC230100027).